\def\year{2022}\relax
\documentclass[letterpaper]{article} 
\usepackage{aaai22}  
\usepackage{times}  
\usepackage{helvet}  
\usepackage{courier}  
\usepackage[hyphens]{url}  
\usepackage{graphicx} 
\usepackage{natbib}  
\usepackage{caption} 
\usepackage{xcolor}
\DeclareCaptionStyle{ruled}{labelfont=normalfont,labelsep=colon,strut=off} 
\usepackage{algorithm}
\usepackage{algorithmic}
\usepackage[colorlinks=true,allcolors=blue]{hyperref}
\usepackage{newfloat}
\usepackage{listings}
\floatstyle{ruled}
\newfloat{listing}{tb}{lst}{}
\floatname{listing}{Listing}
\nocopyright
\usepackage{amssymb}
\usepackage{booktabs}
\usepackage{multirow}
\usepackage{nicematrix}
\usepackage{flushend}
\usepackage[export]{adjustbox}

\title{Exploring Multi-Modal Representations for Ambiguity Detection \& Coreference Resolution in the SIMMC 2.0 Challenge}

\author{
    Javier Chiyah-Garcia\textsuperscript{\rm 1},
    Alessandro Suglia\textsuperscript{\rm 1},
    José Lopes\textsuperscript{\rm 2},\\
    Arash Eshghi\textsuperscript{\rm 1},
    Helen Hastie \textsuperscript{\rm 1}
}
\affiliations{

    \textsuperscript{\rm 1}Heriot-Watt University, Edinburgh, United Kingdom\\
    \textsuperscript{\rm 2} Semasio, Porto, Portugal\\
    \{fjc3, a.suglia, a.eshghi, h.hastie\}@hw.ac.uk, jose@semasio.com
}

\ifdefined\paperDraft
 \def\JCGcomment#1{{\color{teal}[Javier: \textit{#1}]}}
 \def\AEcomment#1{{\color{purple}[Arash: \textit{#1}]}}
 \def\AScomment#1{{\color{red}[Alessandro: \textit{#1}]}}
  \def\HHcomment#1{{\color{orange}[Helen: \textit{#1}]}}
 \def\JCGedit#1{{\color{teal} #1}}
 
 \def\ASedit#1{{\color{red} #1}}
 \def\HHedit#1{{\color{orange} #1}}
\else
 \def\JCGcomment#1{}
 \def\JCGedit#1{{#1}}
 \def\AEcomment#1{}
 
  \def\AScomment#1{}
 \def\ASedit#1{{#1}}
 \def\HHcomment#1{}
 \def\HHedit#1{{#1}}
\fi

\begin{document}

\maketitle


\begin{abstract}
Anaphoric expressions, such as pronouns and referential descriptions, are situated with respect to the linguistic context of prior turns, as well as, 
the immediate visual environment. 
However, a speaker's referential descriptions do not always uniquely identify the referent, leading to ambiguities in need of resolution through subsequent clarificational exchanges. Thus, effective \emph{Ambiguity Detection} and \emph{Coreference Resolution} are key to task success in Conversational AI. In this paper, we present models for these two tasks as part of the SIMMC 2.0 Challenge \citep{kottur-etal-2021-simmc}. Specifically, we use TOD-BERT and LXMERT based models, compare them to a number of baselines and provide ablation experiments.
Our results show that (1) language models are able to exploit correlations in the data to detect ambiguity; and (2) unimodal coreference resolution models can avoid the need for a vision component, through the use of smart object representations.

\end{abstract}


\section{Introduction}

Situated dialogues that involve performing a task in the real world, by their very nature, leverage multi-modal context including language and visual understanding of the scene~\cite{bisk2020experience}. When developing conversational agents to perform a role in the real world, it is important to understand what multi-modal data is needed in order to converse in an effective manner\JCGedit{. This includes} deciding when to request clarification in order disambiguate and performing coreference resolution. 

In this paper, we present our submission to the SIMMC 2.0  \cite{kottur-etal-2021-simmc} challenge where the goal is to create a conversational agent who acts as a shopping assistant. As with human shopping assistants, they need to converse with customers and provide help and respond to questions, based on metadata only they have access to, and within the context of a joint view of the world, grounded in an immersive scene.

The main contributions of this paper are twofold. Firstly, we describe a model for disambiguation prediction using TOD-BERT \cite{Wu.etal19} and a model for coreference resolution that uses dialogue context and object descriptions based on LXMERT \cite{Tan.Bansal19}. Secondly, we present an extensive empirical analysis and discussion of the models, the sub-tasks and the challenge data including exploring the various multi-modal representations available to the model. We
provide experimentation that exploits
natural biases in the data and shows an increase in performance. We release the models, scripts to run experiments and additional analysis with this paper\footnote{\href{https://github.com/jchiyah/exploring-mm-in-simmc2}{https://github.com/jchiyah/exploring-mm-in-simmc2}}.

\section{The SIMMC 2.0 Dataset}

The Situated Interactive MultiModal Conversations (SIMMC) 2.0 dataset \cite{kottur-etal-2021-simmc} is a collection of task-oriented dialogues in a multi-modal setting, where both the system and the agent are situated in the same virtual environment. The dialogues are in the domains of fashion and furniture, and set in a virtual shop where the system acts as the shopping assistant. There are a total of 11,244 dialogues, collected through a combination of self-play and human paraphrase annotation, and randomly split into train / dev / devtest / test-std with 65\% / 5\% / 15\% / 15\% respectively. The test-std split is used as the main evaluation of the final challenge results, but the prediction targets (gold data) is not publicly released. See \citet{kottur-etal-2021-simmc} for more details.

The corresponding challenge contains four sub-tasks: disambiguation, coreference resolution, dialogue state tracking and response generation and retrieval. Our team submitted models to the first two sub-tasks, which we explain below. 

\subsection{Multi-Modal Coreference Resolution and Disambiguation}


Language use in dialogue is highly context dependent. The meaning of what is said not only depends on the context 
provided by previous conversational turns; it can also depend on the objects, events, or people in the immediate environment of a conversation. There are many context-dependent phenomena in dialogue, such as fragments, ellipsis and anaphoric expressions. To interpret anaphoric expressions, such as pronouns and definite referential descriptions, their antecedents need to be identified in the context of the dialogue. This context could be either in the dialogue history, or cross-modally in the non-linguistic, visual context of the environment.
This could mean resolving a pronoun to its antecedent (e.g., ``it'' as a pronoun for coat in ``\textit{Bring the coat please. \underline{It} is on the chair}'') or to a physical/virtual entity in the non-linguistic context (e.g., ``\textit{Bring the blue jacket}''). 

Using the dialogue history and looking at previously mentioned entities may, on occasion, be enough (e.g., ``it'' for coat); alternatively it may be necessary to look at the image of the scene and refer to one of the objects there (e.g., ``\textit{the blue jacket}''). Furthermore, it may be necessary to employ both simultaneously and use both the context and mentioned object properties to discriminate against alternative objects.

\paragraph{Multi-Modal Coreference Resolution (Sub-Task \#2)} For this sub-task, the model has to predict a list of object IDs in the image, which could be empty if there are no references in the current user utterance, or contain more than one for multiple objects. The model can use the current utterance, the dialogue history up to that point and an image of the scene. There is also scene information including a list of the visible objects along with their bounding boxes and the object IDs (prediction targets). These objects can be linked to 
metadata with visual and non-visual properties (i.e., colour and price respectively), although we cannot use any visual metadata at inference time.
The evaluation metric for this sub-task is \textit{object F1}: the mean of precision and recall when comparing the true list with the predicted list of object IDs.

\paragraph{Multi-Modal Disambiguation (Sub-Task \#1)} In dialogue, speakers produce their turns spontaneously in real-time, often in dynamic environments, and where there is potential mismatch between the perspectives of the speaker and hearer (see e.g. \citet{Dobnik.etal15}). This can lead to referring expressions that do not necessarily identify the referent uniquely. \JCGedit{This could lead} to referential ambiguities, which need to be resolved through subsequent clarificational exchanges \cite{Purver04} for repairing the miscommunication \cite{Purver.etal18}. However to do this, hearers need to identify ambiguities first.

For the SIMMC 2.0 Multi-Modal Disambiguation Task, our model has to detect ambiguities in the dialogue. Ambiguities may occur for several reasons, such as when the user does not give enough details to single out a unique referent (e.g., ``\textit{What is the price of that t-shirt?}'' instead of ``\textit{What is the price of the t-shirt on the left?}''). 
Formally, for this task, we need to predict a binary label for whether the system needs to disambiguate or not in the next turn (i.e. ask a clarification question) given the current user utterance, the dialogue history up to that point and an image of the scene.  Metadata and dialogue information is also available as input to the model. As this can be treated as a binary classification task,  \emph{accuracy} is used as the evaluation metric.



\subsection{Baselines}

\citet{kottur-etal-2021-simmc} provide baselines for both sub-tasks with fine-tuned GPT-2-based models \cite{radford2019language}. For Sub-Task \#1, the GPT-2 model uses a classification head to predict the binary disambiguation value from the current user utterance and up to 5 previous utterances from the dialogue history. It achieves 73.9\% accuracy.

The model for Sub-Task \#2 is also the baseline provided for Sub-Tasks \#3 and \#4 of the challenge. In a generative fashion, it predicts the object IDs using special multi-modal tokens \texttt{\textless SOM\textgreater} and \texttt{\textless EOM\textgreater} from the user utterance and dialogue history. 
Optionally, and used by default, the model uses the previously mentioned object IDs in the input with the special tokens. They report an object F1 of 36.6\%. 

We compare both of these baselines against our models in Tables \ref{tab:subtask1-summary} and \ref{tab:subtask2-summary}. We also include our baseline replications.

\section{Models}

We submitted two different models, one for each sub-task. Both are transformer models based on BERT \cite{Devlin2019} and use similar inputs. However, we do not use the multi-modal information in Sub-Task \#1. We report our model metrics of the best performance on dev and then evaluated in devtest dataset splits, unless otherwise stated. The final challenge results, however, were from evaluations on the test-std split, but this is not publicly available. 


\subsection{Sub-Task \#1 Model}

We adapted TOD-BERT \citep{Wu.etal19}, a common pre-trained Transformer-based model for task-oriented dialogues. This model is initialised from BERT-base \cite{Devlin2019} with 12 layers and 12 attention heads, and then trained on task-oriented datasets with masked language modelling and response contrastive losses. Each utterance in the dialogue starts with a special token (i.e., [SYS] for system and [USR] for user) and all turns are concatenated into a flat sequence. Following \citet{Wu.etal19}, a [CLS] token is added to the dialogue representation to obtain a global representation of the dialogue.

For Sub-Task \#1, we modify the SIMMC 2.0 data and give it to TOD-BERT as a binary classification problem, where the model needs to predict whether to disambiguate or not at a given turn. We use the dialogue history up to that point including the current user utterance. The output representation is the final [CLS] embedding, which is passed through a Sigmoid layer to obtain the binary result.

\[ D_{label} = Sigmoid(M(H)) \in \{0, 1\} \]

where $H$ is the dialogue history as the concatenation of utterances ``\textit{[CLS] [SYS] $S_1$ [USR] $U_1$ ... [SYS] $S_t$ [USR] $U_t$}'' up to turn $t$ and $M$ our proposed TOD-BERT model. We train using Binary Cross-Entropy (BCE) loss for 10 epochs, and our final submission uses the pre-trained TOD-BERT-jnt version. Our results for this task are presented in Table~\ref{tab:subtask1-summary}.


\begin{table}[ht]
\centering
\begin{tabular}{clcc}
& \multirow{2}{*}{\textbf{Models for Sub-Task \#1}}                     & \multirow{2}{*}{\textbf{Acc}}   & \textbf{History}   \\
&                                                                       &                                      & \textbf{Turns}     \\ \toprule

\multirow{12}{*}{\rotatebox[origin=c]{90}{devtest}}
& Baseline Random  & 50.1                 & 0                               \\ 
\cmidrule(lr){2-4}

& Baseline \cite{kottur-etal-2021-simmc}               & 73.9       & 5                               \\ 
& Our replication of baseline & 71.0               & 5                               \\ 
& Our replication of baseline &                     &            \\ 
& with hyperparameter tuning                       & 88.0               & 5                               \\ \cmidrule(lr){2-4}
& TOD-BERT BERT-base                               & 91.5               & all                             \\ 
& TOD-BERT BERT-large                              & 89.6               & 5                               \\ 
& TOD-BERT GPT-2                                    & 91.3               & 0                              \\ 
& DialoGPT$_{m}$ \cite{zhang2019-DialoGPTLargeScaleGenerative}                        & 91.3               & 5                               \\ 
& ROBERTA$_{base}$ \cite{liu2019roberta}                    & 90.3               & 5                               \\ 
& LXMERT \cite{Tan.Bansal19}                               & 89.4               & 0                               \\ \cmidrule(lr){2-4}
& Our submission                       & \textbf{91.9}      & 0                               \\ \midrule

\multirow{3}{*}{\rotatebox[origin=c]{90}{test-std}} &
\textbf{Final Challenge Results}                       &       &                                \\ \cmidrule(lr){2-4}
& Our submission                       & 93.1      & 0                               \\ 
& Sub-Task Winner                                  &  94.7     & -                                \\ \bottomrule
\end{tabular}
\caption{\JCGcomment{should I remove the turn column?} \AScomment{I personally believe this is not required. Are we trying to say that we don't really need the history for this task? If that's the case, then let's keep it!} \JCGcomment{I thought that adding the turns column shows that dialogue history isn't needed, for discussion in section 4} Comparison of our entry to the Sub-Task \#1 of the challenge with the baseline provided and several other models. `History Turns' refers to the dialogue history used in the model, i.e., the number of previous utterances given from both system and user. When this is 0, \JCGedit{we do not concatenate any previous dialogue history and} only the current user utterance is used.} 
\label{tab:subtask1-summary}
\end{table}

\subsection{Sub-Task \#2 Model}\label{sec:subtask2-model}

For this sub-task, we extracted visual features to use in our multi-modal model. We will first explain this visual extraction, followed by the model definition.

\paragraph{Visual features extraction}

Our visual feature extraction module is based on Detectron2 \citep{wu2019detectron2} and pre-trained on MS COCO \citep{Lin.etal14}. 
We derive object class labels by combining both object colour and type available in the metadata files. We simplify some colours (e.g., \textit{light blue} becomes \textit{blue}) and depending on the dialogue domain (fashion or furniture), we use \texttt{type} or \texttt{assetType} fields of the objects. In the end, there are 17 colours and 21 object types, with a total of 169 combinations appearing in the data. This will represent the overall set of object classes. 

We fine-tune this model using a multi-task objective that jointly optimises object classification as well as bounding box coordinates prediction, following the training regime presented in \citet{wu2019detectron2}. 



Although it is permitted to use the gold bounding boxes at test time for the model in Sub-Task \#2, there was a considerable improvement in object colour and type accuracy  when we included bounding box prediction loss in the training (from 41.3\% to 65.4\% mean accuracy on the devtest). It seems that predicting bounding boxes helped the model learn where and what salient object features it should focus on. 
At evaluation and inference time, we give the model the gold bounding boxes to force a class prediction in that specific region of the image.
We also modify the model to expose the final layers and extract the Region of Interest (RoI) features to use in our vision encoder.

We use the same dataset splits as in the original SIMMC 2.0 data, however, for images we only train on those that were referenced in the dialogues in the train set, and then evaluate on the dev set. The final results are reported on devtest images. Once the model is fine-tuned, we extract the RoI features and classes with colours and object types of all the images in the SIMMC 2.0 data.

\paragraph{Multi-Modal model definition}
We used an LXMERT \citep{Tan.Bansal19} model as the encoder for our submission in Sub-Task \#2. LXMERT is a Transformer-based model pre-trained using a large captions dataset. It is a dual-stream Transformer encoder, which first encodes vision and language using dedicated encoders, and then uses a cross-modal attention layers to obtain multi-modal contextual representations for visual and text tokens. The final output sequences are the hidden representation of these three encoders, with the cross-modal output being the hidden state of the special token [CLS] appended before the language sequence~\citep{Devlin2019}.

\begin{figure*}[t]
\centering
\includegraphics[width=0.9\textwidth]{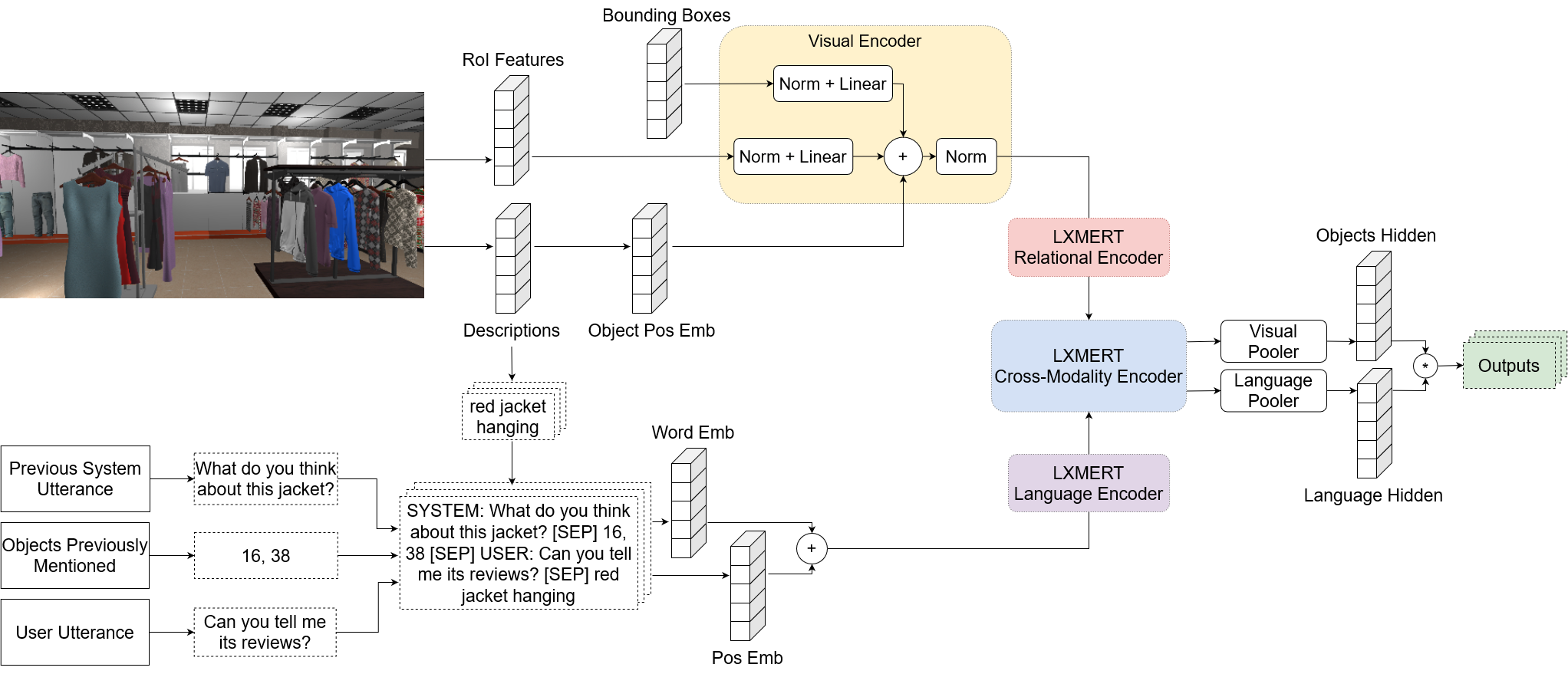}
\caption{The architecture of the model submitted for Sub-Task \#2. It uses vision to derive textual descriptions, and it combines the RoI features with language to obtain the final outputs. `Norm' and `Linear' stand for normalisation and linear layers.}
\label{fig:architecture}
\end{figure*}

For our coreference resolution model, we combine the object colour and type properties into textual descriptions of each object (e.g. ``blue trousers display''), $D = \{d_1, d_2, ..., d_n\}$. We then assemble a natural language sentence with the user utterance $U_t$ at turn $t$, for each one of these descriptions, separated by a special [SEP] token e.g., ``$U_t$ [SEP] $d_1$'', see the bottom part of Fig \ref{fig:architecture}. Next, we append to this sentence the previous system utterance, $S_{t-1}$ and previously mentioned objects by the system\footnote{from the \texttt{system\_transcript\_annotated} field at $t-1$}, $O_{t-x}$, where $x$ is the amount of previous turns to use. We append tokens `SYSTEM' and `USER' to each utterance in $S_{t-1}$ and $U_t$ to maintain speaker information. The final language sentences have the form:

\[l_n = S_{t-1} [SEP] O_{t-x} [SEP] U_t [SEP] d_n \]    

An example language sequence is: 
``\textit{SYSTEM : We have this red, white and yellow blouse on the left, and the white blouse on the right [SEP] 60 56 [SEP] USER : I'd like to get that blouse, please [SEP] white blouse hanging}''. Similar to the tasks of sentence entailment or QA \cite{Devlin2019}, we denote everything in the sequence up to the last [SEP] as the context, and $d_n$ as the question, with segment IDs.

These text sequences, $L_t = \{l_1, ..., l_n\}$, are fed to the model along with the corresponding object's visual RoI features, bounding boxes and object positional ID. The object positions represent the amount of objects per type in the scene and are derived from the object types similar to how word position IDs work (e.g., the first jacket will be 1, the second one will be 2, etc.). These positional IDs are used to help the model better recognise different objects that have the same type (and potentially the same colour). 

The model first extracts BERT embeddings from $L_t$, as well as object positional embeddings from the object positions. The visual encoder combines the bounding box positions, $p_n$, RoI features $f_n$ and object positional embeddings $e_n$ through several linear, and normalisation layers to obtain a visual feature state $v_n$, as shown at the top part of Fig \ref{fig:architecture}.

\[ p_n = LayerNorm ( W_p p_n + b_p) \]
\[ f_n = LayerNorm ( W_f f_n + b_f) \]
\[ e_n = PositionEmb ( W_e e_n + b_e) \]
\[ v_n = Dropout ( LayerNorm ( p_n + f_n + e_n ) ) ) \]


The sentence embeddings and visual features are passed through the typical set of LXMERT encoders (language, relational and cross-modality layers) to obtain two feature sequences with shared modality information: language and vision $h^l_n$ and $h^v_n$. We then apply pooling and compute the dot product to extract a hidden cross-modal representation $X_n$ that we pass through a sequence of simple GeLU, normalisation and fully connected linear layers to obtain our output vector.

\[ h^l_n, h^v_n = LXMERT^{encoder} ( l_n, v_n ) \]

\[ X_n = LayerPool ( h^l_n) \cdot LayerPool ( h^v_n ) \]


The prediction target for Sub-Task \#2 is a list of object indexes for those that were referenced at $t$. 
Therefore, we treat the task as a multi-label classification problem where output is a vector of $|L_t|$ binary values, one for each object given in input to the model. The model output is defined as follows:

\[ prob^N_{n,t} = Sigmoid ( W \cdot LayerNorm ( GeLU ( X_n ) ) + b ) \]


where $prob(n, t)$ refers to the probability of object $n$ referenced at turn $t$. The predicted object IDs are selected when their corresponding score is above a threshold\footnote{We used the validation set to define the best threshold value for each model. Usually this is a value between $0.3$ and $0.4$.}. We minimise the BCE loss during training for each $l_n$ separately and compute the mean ignoring padded values. 






Due to size limitations, we reduce the number of layers used in the original LXMERT encoders from 9 in language, 5 relational and 5 cross-modality to 5, 3 and 3 layers respectively. We also remove from training all scenes with more than 60 objects, equal to 4\% of the train data, to reduce the maximum size of $N$. Scenes have a mean of $27.6 \pm 20.7$ objects, with some of them reaching up to a maximum of 141. We keep these scenes in the data at evaluation and test time and increase $N$ as needed. We train for a total of 10 epochs and our submitted model achieves 57.3\% object F1 in the final challenge results. Table \ref{tab:subtask2-summary} shows a comparison of our model with the baseline and winner of the Sub-Task. See Table \ref{tab:subtask2-ablation} for model ablations, which we will discuss in Section \ref{sec:subtask2-experiments}.

\begin{table}[ht]
\centering
\begin{tabular}{c l c}
& \textbf{Models for Sub-Task \#2}                    & \textbf{Object F1 ↑}      \\ \toprule
\multirow{3}{*}{\rotatebox[origin=c]{90}{devtest}} 

& Baseline \cite{kottur-etal-2021-simmc}              & 36.6                      \\ 
& Our replication of baseline                         & 36.4                      \\ 
& Our submission                                      & 57.4                      \\ \midrule

\multirow{3}{*}{\rotatebox[origin=c]{90}{test-std}}
& \textbf{Final Challenge Results}                    &                           \\ \cmidrule(lr){2-3}
& Our submission                                      & 57.3                      \\ 
& Sub-Task Winner                                     & 75.8                      \\ \bottomrule

\end{tabular}
\caption{Comparison of our entry to the Sub-Task \#2 of the SIMMC 2.0 challenge with the baseline provided and the challenge winner. \JCGcomment{Note that the baseline comparison is done with the devtest split, whereas the final results were on the test-std split, and hence we report different object F1s.}}

\label{tab:subtask2-summary}
\end{table}


A disadvantage of our model architecture is that we are not able to handle coreferences 
involving the same object multiple times. For instance, the user may say ``\textit{USER : I'll take two of the wood one}'' to refer to two of the same item, for the prediction target [7, 7], and we will address this in future work. 








\section{Sub-Task \#1 Experiments}

As seen in Table \ref{tab:subtask1-summary}, models achieve relatively high accuracy in Sub-Task \#1 regardless of their complexity. The provided baseline from \citet{kottur-etal-2021-simmc} reaches 73.9\% accuracy, and we are able to increase its accuracy to 88.03\% by changing the batch size and removing any dialogue history. In this section, we explore the data and models, and discuss possible reasons for this increase in accuracy.

The data is evenly balanced between the disambiguation labels in train, dev and devtest (devtest has disambiguate True 49.4\% and False 50.6\%). \JCGcomment{citation needed} A need to disambiguate often arises after mismatches in expectations from one of the conversation participants, and the dialogue history might indicate such situations. It could also be used to indicate the number of discourse entities, with more entities being a signal for disambiguation. However, including turns before the user utterance that trigger (or not) the disambiguation do not seem to increase model performance. 
We show that our submission, TOD-BERT, as well as other architectures that do not use dialogue history (LXMERT) perform competitively with other models that use more context. In our experiments, dialogue history is optional and models learn faster and sometimes better with fewer history utterances. In fact, none of our models in Table \ref{tab:subtask1-summary} use multi-modal input \JCGcomment{for this Sub-Task \#1}, including the LXMERT model (RoI features were \JCGcomment{masked as} zeros).

We hypothesise that the unimodal models are exploiting correlations in the data between particular syntactic structures and the disambiguation label, e.g. more definite references, more pronoun usage, etc. We therefore first perform Part of Speech (POS) tagging using SpaCy \cite{Honnibal_spaCy_Industrial-strength_Natural_2020}. Our analyses show that the utterances with the two disambiguation labels (\texttt{disambiguation / no disambiguation}) have similar POS \JCGcomment{tag} distributions, with prepositions and adjectives slightly more common in cases of \texttt{no disambiguation}. A Pearson coefficient between the number of prepositions in a sentence and \texttt{no disambiguation} shows a medium-weak \HHedit{but significant} correlation ($r=0.402$, $p=9.47\mathrm{e}{-75}$).\JCGcomment{Should I also report degrees of freedom for Pearson's R?} Adjectives also show a similar weak \HHedit{but significant} correlation ($r=0.291$, $p=2.27\mathrm{e}{-38}$). In other words, there seems to be some negative relationship between the need for disambiguation and the number of prepositional expressions 
(e.g., ``\textit{in the top shelf}'') and adjectives (e.g., ``\textit{light blue jacket}''). For example, ``\textit{Can you tell me the sizes and prices \underline{of} the two \textbf{grey} jeans \underline{in} the \textbf{back} two cubbies \underline{on} the right?}'', where \underline{\textit{underline}} denotes prepositions and \textbf{\textit{bold}} words adjectives. 





On the other hand, a Pearson coefficient between the number of times a wh-question (\textit{what, who}) appears and disambiguation shows a medium-weak \HHedit{significant} correlation ($r=0.323$, p=$2.22\mathrm{e}{-47}$). Thus, user utterances of the form ``\textit{What's the price on those two rugs?}'' or ``\textit{Can you tell me who makes that sofa chair on the right?}'' are more likely to need disambiguation. We also observe many other weak correlations in the data with other type of tags. Combined with the mean sentence length difference ($13.94\pm 6.34$ for the \texttt{disambiguation} label compared to $18.66\pm 8.79$ for the \texttt{no disambiguation label}), it suggests that a powerful language model may be able to exploit these biases and successfully predict the majority of the utterances without using dialogue history or multi-modal input.

\section{Sub-Task \#2 Experiments}\label{sec:subtask2-experiments}

We divide this section in two parts. First, we explore our model with ablations of the different components and discuss what worked and why some parts did not; and second, we run experiments on the data with our model and baselines to further analyse the challenge dataset. We find that multi-modality is not needed to achieve high performance.  

\subsection{Model Ablations}

Table \ref{tab:subtask2-ablation} shows ablations of our model for Sub-Task \#2. We use a refined version of our submitted model with better hyperparameters and we mask with 0s the ablated input features. All model ablations are trained with the same hyperparameters for 10 epochs, and evaluated on the devtest split. We discover that language is the most important feature.

\begin{table}[ht]
\centering
\begin{tabular}{r l c c}
& \textbf{Coreference Model Ablations}                                                        & \textbf{F1 ↑} & \textbf{$\Delta$} \\ \toprule

1. & Refined Full Model                                                                          & 60.83                & -                           \\ \midrule
2. & Baseline \cite{kottur-etal-2021-simmc}                                                      & 36.37                & -24.46                      \\ 
3. & \hspace{1ex}no Object IDs                                                                   & 15.23                & -45.60                      \\ 

\midrule

 & \textbf{Vision}                                                                             &                      &                             \\
5. & \hspace{1ex}no RoI features                                                                 & 57.44                & -3.39                           \\
6. & \hspace{1ex}no bounding boxes                                                               & 60.83                & 0                           \\ 
7. & \hspace{1ex}no object position counts                                                       & 60.78                & -0.05                        \\
8. & \hspace{1ex}no visual or cross-modal enc                                                    & 57.64                & -3.19                        \\
\midrule

    & \textbf{Language}                                                                           &                      &                             \\
9.  & \hspace{1ex}no user utterances                                                              & 33.72                & -27.11                      \\

10. & \hspace{1ex}no previous system turn                                                         & 60.74                & -0.09                       \\
11. & \hspace{1ex}no previous objects                                                             & 48.93                & -11.9                       \\ 

12. & \hspace{1ex}no descriptions                                                                 & 54.50                & -6.33                      \\
13. & \hspace{1ex}no colours in descriptions                                                      & 57.36                & -3.47                       \\
14. & \hspace{1ex}no types in descriptions                                                        & 57.52                & -3.31                       \\
15. & \hspace{1ex}no object IDs in descriptions                                                   & 48.98                & -11.85                      \\
\midrule
16. & with oracle descriptions (upper)                                                            & \textbf{70.71}                &  +9.88                      \\ \bottomrule

\end{tabular}
\caption{Ablations for our model in Sub-Task \#2 compared with the baseline. The difference, Delta $\Delta$, is in respect to the full model (Row 1).\JCGcomment{This refined model has the same architecture as the submitted model, and we only tweak hyperparameters such as batch size, gradient accumulation and learning rate. We evaluate two different versions of the baseline, with and without multi-modal input in rows 2 and 3 respectively. We mask with 0s the ablated input features.}}
\label{tab:subtask2-ablation}
\end{table}


Removing vision-related components decreases the model's performance slightly. RoI features seem the most important in this category, with object F1 decreasing 3.39\% compared to the full model (row 5 compared to row 1). Ablating bounding boxes does not modify the performance (row 6), likely because they might be too noisy in most scenes and not provide enough fine-grained details to discriminate objects. This is a clear requirement for this dataset where there are multiple objects of the same class to choose from. The object position counts increase the object F1 by 0.05\%, likely in scenes with many similar objects, but they rely on the descriptions from Detectron2 and hence may be inaccurate. Finally, in row 8, we fully remove the connections to the visual, relational and cross-modal encoders of LXMERT and use the the language encoder only. We also skip the language pooling layer and use a fully connected layer to obtain an ablation comparable with the others.
We further explore the RoI features by computing their similarity scores across objects with similar descriptions. Their mean is 0.62 (SD: 0.14), which gives us an idea that these features are generally similar to each other and therefore the model may have a hard time spotting the correct object. Additionally, we use t-SNE~\cite{van2008visualizing} to project the representations in a shared embeddings space and observed 
that they are mostly clustered by class type, which can hinder performance when it comes to discriminating the target object in a very cluttered scene.

Regarding language, user utterances seem the most important feature, showing that our model correctly makes use of what the user says to find the referenced object. However, the previous system turn does not seem very helpful and only increases object F1 slightly. The previously mentioned objects, a list of the object IDs that the system mentioned in the last turn, seem to be key features, as a model without these performs poorly with an absolute loss of -11.9\% object F1. 

As discussed in Section \ref{sec:subtask2-model}, the descriptions are natural language sequences derived from the classes extracted with our Detectron2 model. In some way, these represent the human-readable encoding of the RoI features, as they both come from the same vision model. Unlike with the vision ablation, the coreference model seems to learn how to use these descriptions to identify objects and thus increase the object F1. They provide a way to discriminate objects with different properties (e.g., ``\textit{blue jacket hanging}'' vs ``\textit{red jacket hanging}''), likely very useful in scenes with many similar objects where only a few properties distinguish each other. Indeed, there are 21 types of objects and 17 colours across both furniture and fashion domains, but each scene has a mean of $27.6 \pm 20.68$ objects. In fact, removing colours from the descriptions decreases the object F1 more than removing types (rows 13 and 14 in Table \ref{tab:subtask2-ablation}). Providing gold descriptions increases the performance by 9.88\%, suggesting that the model is far from reaching its full potential. We include the canonical object IDs in the descriptions and this also seems crucial for the model's performance. We believe that using these IDs provides the model with a strong signal and that they act in a complementary manner to the object properties. These IDs can be related to the previously mentioned object IDs, which may provide a way to exchange information with the context and further reinforce signals to correctly predict the referenced object.

For the GPT-2 baseline, we also provide an ablation without the multi-modal context in row 3. We remove the previously mentioned object IDs from the input (e.g., \texttt{\textless SOM\textgreater 56 32\textless EOM\textgreater}) and reach an object F1 of 15.23\%, compared to 36.37\% from the full baseline. There are two items for discussion with regards this baseline. 

Firstly, neither version of the baseline model has a notion of the number of objects or the IDs of these in the scene. Unlike our model, aware of the total number of objects $N$, the baseline model only uses the dialogue history, user utterance and \JCGcomment{previously} mentioned objects \JCGcomment{by the system} to predict the canonical object IDs. These IDs are pseudo-randomised between scenes and they do not sequentially start at 0 or 1, so an object F1 of 15.23\% seems fairly high. 

Secondly, the full baseline model seems to take advantage of the previously mentioned object IDs by learning how to relate them to the dialogue and/or user utterance. This should be difficult as the IDs are not unique, e.g., ID 0 could be a red jacket in one scene but a wooden table in another one. However, only 51.6\% of the predicted object IDs appear in the previously mentioned objects, whereas the other 48.4\% are hallucinated or, more likely, relate to some underlying object representation that the model is able to learn, similar to how the ablated baseline reaches 15.23\%.
We will discuss the impact of using object IDs in the next section.



\subsection{Task Analysis}

In this section, we discuss the task, data and models and investigate emergent properties of these. 

\paragraph{Object IDs} Some ablations in the previous section show that the object IDs are helping models learn and solve the task, thus we need to understand how they appear throughout the data. Each object in a scene comes with a list of properties, such as the bounding boxes of the object in the image, a prefab model ID and the \texttt{object\_index} property that we use for the canonical object IDs. The prefab model IDs are used to relate an object to its metadata properties (e.g., colour, type, price, size...) and objects that share the same model ID are exactly the same in every scene. 

On the other hand, object indexes, or object IDs, are not unique throughout scenes as object IDs may be assigned to different model IDs. Initially, it seems there is not a relation between object IDs and the object properties since they are randomly assigned. Each object ID follows a normal distribution mirroring the objects in the data (e.g., 36\% of the time, object ID 0 relates to a jacket, 18\% as a blouse, etc.).
These object IDs are not necessarily sequential either, they might start at 60 in some scenes, or 0 in others, and skip numbers in between. However, they are skewed towards smaller numbers, particularly the target IDs. In fact, 52.5\% of the coreference prediction targets have an object ID below 10, and 30.1\% below 5, suggesting that models can learn to give more weight to smaller IDs and ignore the densely populated scenes altogether.



\paragraph{Exploiting object information} It is possible to exploit the information available in the scenes to further increase the model performance. As mentioned previously, the scenes have the prefab model ID for each object, which is unique and relates to the metadata files. Although we are not allowed to use parts of the metadata, everything in the scene information is usable at evaluation time. Therefore, we could create an object representation from these model IDs that carries across scenes and a model is able to relate to object properties mentioned in the text. 

For this experiment, we generate special global tokens to represent each model ID (i.e., model ID \texttt{[...]/Lamp\_CHLH3796} becomes \texttt{[OBJECT\_320]}) and use this instead of object IDs. In some way, this is amplifying the connection that models exploit with the object IDs by allowing them to create a more robust latent object representation that they can use in all splits of the data. Results are shown at the bottom of Table \ref{tab:subtask2-experiments}.

These representations improve the model's performance from 60.83\% to 68.53\% (+7.7\%), suggesting that the model is able to leverage these tokens to learn additional properties. The model's object F1 increases when we remove the descriptions (row 10 in Table \ref{sec:subtask2-experiments}), likely due to their inaccuracy or noise. However, an oracle upper bound of these descriptions (if the Detectron2 model had perfect accuracy) shows that the model could push the object F1 even higher to 70.53\% and that they could still provide vital information.

\newcounter{rowcount}
\setcounter{rowcount}{0}
\newcommand{\rownumber}{\stepcounter{rowcount}\therowcount.}

\begin{table}[ht]
\centering
\begin{tabular}{r l c c}
           & \textbf{Coreference Model Experiments}                                 & \textbf{F1 ↑}        \\ \toprule

\rownumber & Refined Model                                                          & 60.83                \\ 
\rownumber & \hspace{1ex}with all previous objects                                  & 63.76                \\
\rownumber & \hspace{1ex}no duplicated objects                                      & 71.69                \\
\JCGcomment{I removed \ldots to maintain consistency}
\midrule

           & \textbf{Additional Baselines}                                          &                      \\
\rownumber & Baseline Random (IDs below 141)                                        & 0.65                 \\
\rownumber & Baseline Random (IDs below 5)                                          & 4.69                 \\
\rownumber & Baseline previous objects                                              & 33.16                \\
\rownumber & Baseline all previous objects                                          & 32.15                \\
\midrule

           & \textbf{Global Object Representation}                                  &                      \\
\rownumber & Default (same parameters as row 1)                                     & 68.53                \\
\rownumber & \hspace{1ex}with all previous objects                                  & 68.94                \\
\rownumber & \hspace{1ex}no descriptions                                            & 70.29                \\
\rownumber & \hspace{1ex}no RoI features                                            & 66.72                \\
\rownumber & \hspace{1ex}oracle descriptions (upper bound)                          & \textbf{70.53}                \\
\bottomrule

\end{tabular}
\caption{Experiments for Sub-Task \#2 with different configurations. All models are evaluated on the devtest split. Row 2 is our model with all objects previously mentioned by the system for $O_{0, ..., t}$, instead of $O_{t-1}$. In row 3, we remove objects with duplicated descriptions. For the random baselines, we randomly select an object ID below that number. In rows 6 and 7, we evaluate object F1 with a list of the objects previously mentioned by the system at $t-1$ and all turns respectively. The bottom of the table has the results of using global object tokens based on model IDs.}
\label{tab:subtask2-experiments}
\end{table}


\paragraph{Dialogue history and previously mentioned objects} At inference time, we can access and use the dialogue history and, in particular, a list of the object IDs that the system mentioned per turn. One of the additional baselines that we provide in Table \ref{tab:subtask2-experiments} (row 6) is the result of using the list of objects that the system mentioned in $t-1$ as the predicted IDs. The performance is just below the GPT-2 baseline at 33.16\%, which shows that the referenced objects are often part of the dialogue. Further analysis reveals that the system mentions the target objects 37\% of the time in the previous turns of the dialogue, but a baseline using all of the previously mentioned objects performs slightly worse in practice at 32.15\% (row 7). However, our model benefits from using a list of all objects mentioned so far ($O_{0, ..., t}$), with a net increase of 2.93\% in object F1 over our model (row 2). We see little improvement of using dialogue history in our model, as shown in Table \ref{tab:subtask2-ablation}. In general, this suggests that the coreference targets have to be resolved with a mix of looking at the objects mentioned throughout the conversation, the current user utterance and vision, but not necessarily what the system has previously said.

\paragraph{Ambiguous coreferences} 
The data used in Sub-Task \#1 for disambiguation is a subset of all the data used in Sub-Task \#2, since only a few turns (usually 1 or 2 per dialogue) contain the disambiguation label to predict. Thus, out of 8609 turns in devtest, 1904 have the disambiguation label (22\%). \citet{kottur-etal-2021-simmc} mention that they ``\textit{exclude from evaluation the object mentions that are immediately followed by a disambiguation request  (e.g., `How much is the one over there?' $\to$
`Which one do you mean?') as they do not provide sufficient descriptions for resolving those coreferences}''. In other words, turns where disambiguation is true, so the system has to clarify in the next utterance.
However, our findings suggest that there is no difference in object F1 for these turns. Our submitted model achieves 60.13\% object F1 when evaluated only on the turns with a disambiguation label, higher than our reported 57.44\% for the whole data. Surprisingly, we also find that the object F1 for turns immediately followed by a disambiguation request (those with disambiguation label as true) is 60.89\%, whereas for those with disambiguation as false is 59.24\%.

In the turns after the system asks a clarification question (e.g., ``\textit{Which one do you mean?}'' $\to$ ``\textit{The red ones on the left}''), our model performs much worse at 45.36\%. This is partially explained by our model architecture and the use of the objects mentioned by the system at $turn - 1$. These clarification requests require more history, as people often clarify elliptically, so they do not necessarily repeat the information that was in the previous term and the model loses information. A longer dialogue history and including objects mentioned at $turn-2$ or even $turn-3$ may be able to mitigate this drop in performance.

\JCGcomment{How can I talk about the commented part above/ should I mention anything? Utterances that were originally correct are now not correct }






\JCGcomment{Text to comment out below, but not everything made it to the discussion above}

\section{Conclusion}


Multi-modality is at the centre of the SIMMC 2.0 challenge, yet our results suggest that strong unimodal models can perform well for at least two of the sub-tasks, underlining the importance of unimodal baselines for multimodal tasks~\cite{thomason-etal-2019-shifting}. For ambiguity detection, language models seem to exploit certain biases in the data, such as the appearance of adjectives or the sequence length, to classify between the labels. We show how our disambiguation model and other popular models do not need the dialogue history or the visual context to achieve an accuracy above 90\%.

Regarding coreference resolution, we develop a model based on LXMERT that uses dialogue context and object descriptions extracted from a Detectron2 model, reaching 60.83\% object F1.
The coreference sub-task is more challenging and indeed, it needs both vision and language to achieve top performance. However, the image scenes are sometimes too cluttered with objects, and the object variation may not be large enough to allow vision models to correctly discriminate. \JCGedit{Even descriptions, a noisy textual representation of the RoI features, were more helpful at improving the model's performance.} We can also extract object representations from the data to use in language-only models, bypassing any multi-modality and performing better than with vision. Models are able to learn object representations through the object IDs alone, even if these are randomised and not sequential. The object model IDs can also be used to further increase the performance through a global object representation. \ASedit{This calls for a more systematic way of learning object representations that are disentangled~\cite{locatello2019challenging}. Specifically, this would improve the ability of the model to discriminate the target object among a set of distractors in the same scene.}


%



\section*{Acknowledgements}

Chiyah-Garcia’s PhD is funded under the EPSRC iCase  with Siemens (EP/T517471/1). This work was also supported by the EPSRC CDT in Robotics and Autonomous Systems (EP/L016834/1).  

\bibliography{refs}

\end{document}